\documentclass[10pt,twocolumn,letterpaper]{article}

\usepackage{iccv}
\usepackage{times}
\usepackage{epsfig}
\usepackage{graphicx}
\usepackage{amsmath}
\usepackage{amssymb}
\usepackage{bm}
\usepackage{multirow}
\usepackage{mathrsfs}
\usepackage{floatrow}

\newfloatcommand{capbtabbox}{table}[][\FBwidth]

\usepackage{enumitem}
\setitemize[1]{itemsep=-0.2em,topsep=0em}
\usepackage{subfigure}
\usepackage[font=small]{caption}
\usepackage{setspace}

\usepackage[pagebackref=true,breaklinks=true,letterpaper=true,colorlinks,bookmarks=false]{hyperref}

\iccvfinalcopy 

\def\iccvPaperID{1693} 

\ificcvfinal\fi

\begin{document}

\title{Correlation Congruence for Knowledge Distillation}

\author{Baoyun Peng\thanks{Equal contribution.}\hspace{4pt}\thanks{This work was done when Baoyun Peng was an intern at SenseTime Inc.}\hspace{4pt}$^{1}$, Xiao Jin\footnotemark[1]\hspace{4pt}$^{2}$, Jiaheng Liu$^{3}$, Shunfeng Zhou$^{2}$, Yichao Wu$^{2}$ \\ Yu Liu$^{4}$, Dongsheng Li$^{1}$, Zhaoning Zhang$^{1}$\\
$^{1}$\hspace{2pt}NUDT\hspace{50pt}$^{2}$\hspace{2pt}SenseTime\hspace{50pt}$^{3}$\hspace{2pt}BUAA\hspace{50pt}$^{4}$\hspace{2pt}CUHK\\
{\small \textit{\{pengbaoyun13, dsli, zhangzhaoning\}@nudt.edu.cn}\hspace{20pt} \textit{liujiaheng@buaa.edu.cn}} \\ {\small \textit{\{jinxiao, zhoushunfeng, wuyichao\}@sensetime.com}\hspace{20pt}\textit{yuliu@ee.cuhk.edu.hk}}
}

\maketitle

\begin{abstract}
Most teacher-student frameworks based on knowledge distillation (KD) depend on a strong congruent constraint on instance level. However, they usually ignore the correlation between multiple instances, which is also valuable for knowledge transfer. In this work, we propose a new framework named correlation congruence for knowledge distillation (CCKD), which transfers not only the instance-level information, but also the correlation between instances. Furthermore, a generalized kernel method based on Taylor series expansion is proposed to better capture the correlation between instances. Empirical experiments and ablation studies on image classification tasks (including CIFAR-100, ImageNet-1K) and metric learning tasks (including ReID and Face Recognition) show that the proposed CCKD substantially outperforms the original KD and achieves state-of-the-art accuracy compared with other SOTA KD-based methods. The CCKD can be easily deployed in the majority of the teacher-student framework such as KD and hint-based learning methods. Our code will be released, hoping to nourish our idea to other domains.
\end{abstract}

\section{Introduction}

Over the past few decades, various deep neural network (DNN) models have achieved state-of-the-art performance in many vision tasks \cite{simonyan2014very,szegedy2015going,girshick2014rich}. Generally, networks with many parameters and computations perform superior to those with fewer parameters and computations when trained on the same dataset. Nevertheless, it's difficult to deploy such large networks on resource-limited embedded systems. Along with the increasing demands for low cost networks running on embedded systems, there is an urgency for getting smaller network with less computation and memory consumptions, while narrowing the gap of performance between minor network and large network. 

\begin{figure}[htbp]
    \caption{The difference between instance congruence and correlation congruence. When focusing on only instance congruence, the correlation between instances of student may be much different from the teacher's, and the cohesiveness of intra-class would be worser. CCKD solve the problem by adding a correlation congruence when transferring knowledge. \label{fig:motivation}}
    \begin{center}
    \includegraphics[width=0.98\linewidth]{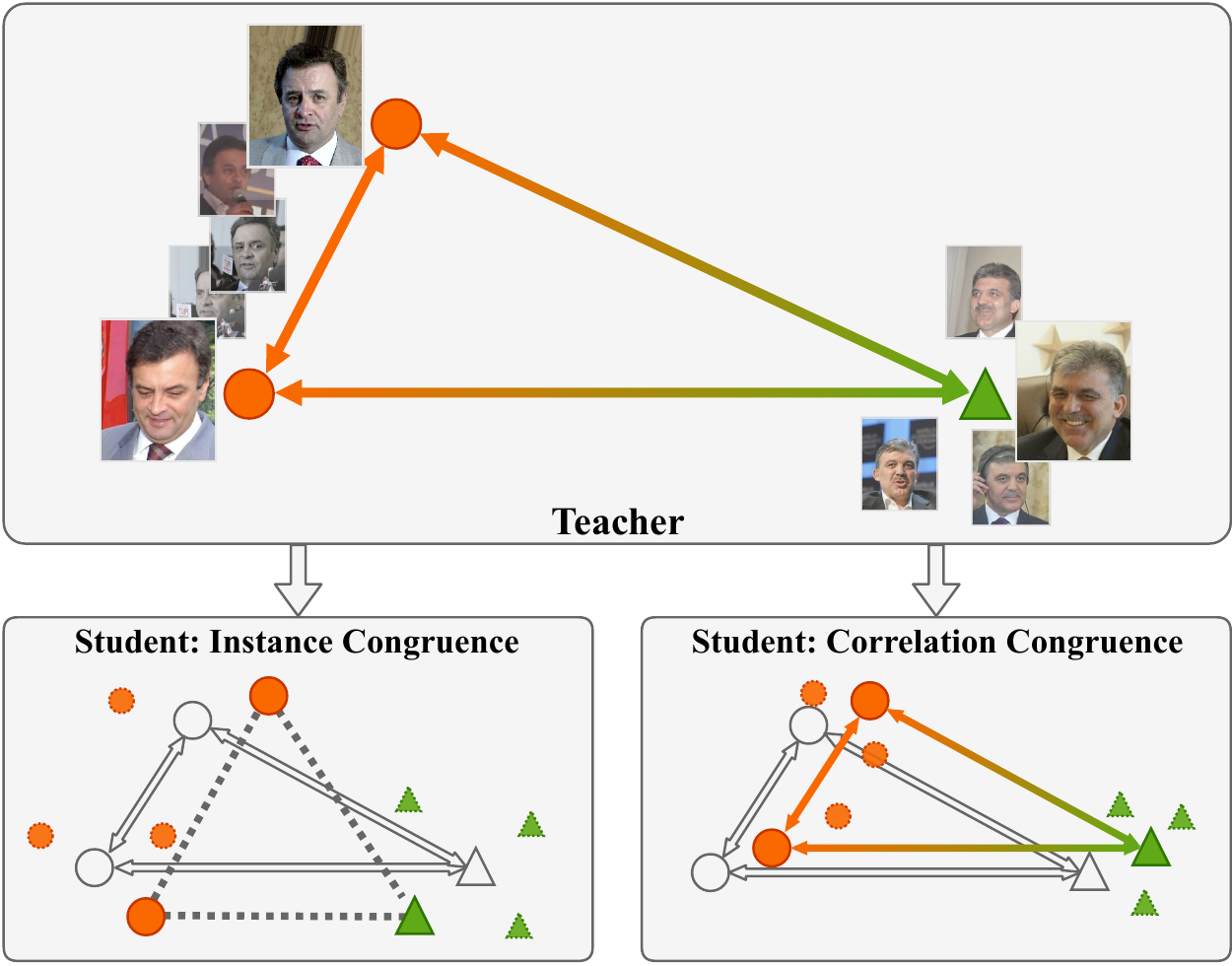}
    \end{center}
   \end{figure}

Several techniques have been proposed to address this issue, e.g. parameter pruning and sharing \cite{han2015learning,Molchanov2016Pruning}, compact convolutional filters \cite{Zhang2017ShuffleNet,howard2017mobilenets}, low-rank factorization \cite{jaderberg2014speeding,denton2014exploiting} and knowledge distillation \cite{hinton2015distilling}. Among these approaches, knowledge distillation has been proved as an effective way to promote the performance of small network by mimicking the behavior of a high-capacity network. It works by adding a strong congruent constraint on outputs of teacher and student for each input instance to encourage the student to mimic teacher's behavior, e.g. minimizing the Kullback–Leibler divergence of predictions \cite{hinton2015distilling} or minimizing the euclidean distance of feature representations \cite{li2017mimicking} between teacher and student.

However, it's hard for the student to learn a mapping function identical to the teacher's due to the gap (in capacity) between teacher and student. By focusing on only instance congruence, the student would learn a much more different instances correlation from the teacher as shown in figure \ref{fig:motivation}. Usually, the embedding space of teacher possesses the characteristic that intra-class instances cohere together while inter-class instances separate from each other. But its counterpart of student model trained by instance congruence would lack such desired characteristic. 

We claim that beyond instance congruence, the correlation between instances is also valuable knowledge for promoting the performance of student. Based on this philosophy, we propose a new distillation framework called Correlation Congruence Knowledge Distillation (CCKD) which focus on not only instance congruence, but also correlation congruence to transfer the correlation knowledge between instances to the student as shown in Figure \ref{fig:motivation}. CCKD can be easily implemented and trained with mini-batch, and only requires the same dimension of embedding space for teacher and student network. To cope with the dismatch of feature representations of teacher student network on image classification tasks, we apply a fully-connected layer with the same dimension for both teacher and student network. We conduct various experiments on four representative tasks and different networks to validate the effectiveness of the proposed approach.

Our contributions in this paper are summarized as follows:
\begin{enumerate}
    \item We propose a new distillation framework named correlation congruence knowledge distillation (CCKD), which focuses on not only instance congruence but also correlation congruence. To the best of our knowledge, it is the first work to introduce correlation congruence to distillation;
    \item We introduce a general kernel-based method to better capture the correlation between instances in a mini-batch. We have evaluated and analyzed the impact of different correlation metrics on different tasks; 
    \item We explore different sampler strategies for mini-batch training to further improve the correlation knowledge transfer;
    \item Extensive empirical experiments and ablation studies show the effectiveness of proposed method in different tasks (CIFAR-100, ImageNet-1K, person re-identification and face recognition) to improve the distillation performance.
\end{enumerate}

\section{Related Work}

Since this paper focuses on training a small but high performance network based on knowledge distillation, we discuss related works in model compression and acceleration, knowledge distillation in this section. In both areas, there are various approaches have been proposed over the past few years. We summarize them as follows.

\textbf{Model Compression and Acceleration}. Model compression and acceleration aim to create network with few computation and parameters cost meanwhile maintaining high performance. A straight way is to design lightweight but powerful network since the original convolution network has many redundant parameters. For example, depthwise separable convolution is used to replacing standard convolution for building block in \cite{howard2017mobilenets}. Pointwise group convolution and channel shuffle are proposed to reduce the burden of computation while maintaining high accuracy in \cite{Zhang2017ShuffleNet}. Another way is network pruning which boosts the speed of inference by pruning the neurons or filters with low importance based on certain criteria \cite{han2015learning,Molchanov2016Pruning}. In \cite{jaderberg2014speeding,denton2014exploiting}, weights were decomposed through low-rank decomposition to save memory cost. Quantization seeks to use low-precision bits to store model's weights or activation outputs \cite{han2015deep,Hubara2016Quantized,Wu2016Quantized}. 

\textbf{Knowledge Distillation}. Transferring knowledge from a large network to a small network is a classical topic and has drawn much attention in recent years. In \cite{hinton2015distilling}, Hinton \etal propose knowledge distillation (KD), in which the student network was trained by the soft output of an ensemble of teacher networks. Comparing to one-hot label, the output from teacher network contains more information about the fine-grained structure among data, consequently helps the student achieve better performance. Since then, there have been works exploring variants of knowledge distillation. In \cite{ba2014deep}, Ba and Caruana show that the performance of a shallower and wider network trained by KD can approximate to deeper ones. Romero \etal \cite{romero2014fitnets} propose to transfer the knowledge using not only final outputs but also intermediate ones, and add a regressor on intermediate layers to match different size of teacher's and student's outputs. In \cite{Zagoruyko2016Paying}, the authors propose an attention-based method to match the activation-based and gradient-based spatial attention maps. In \cite{yim2017gift}, the flow of solution procedure (FSP) , which is generated by computing the Gram matrix of features across layers, was used for knowledge transfer. To improve the robustness of the student, Sau and Balasubramanian \cite{sau2016deep} perturbe the logits of teacher as a regularization. 

Different from above offline training methods, several works adopts collaboratively training strategy. Deep mutual learning \cite{zhang2018deep} conducts distillation collaboratively for peer student models by learning from each other. Anil \etal \cite{anil2018large} further extend this idea by online distillation of multiples networks. In their work, networks are trained in parallel and the knowledge is shared by using distillation loss to accelerate the training process. 

Besides, there are several works utilizing adversarial method to modeling knowledge transfer between teacher and student \cite{xu2018training,heo2018improving,Byeongho2018Knowledge}. In \cite{xu2018training}, they adopt generative adversarial networks combined with distillation to learn the loss function to better transfer teacher's knowledge to student. Byeongho \etal \cite{Byeongho2018Knowledge} adopt adversarial method to discover adversaial samples supporting decision boundary.

In this paper, beyond instance knowledge, we take the correlation in embedded space between instances as valuable knowledge to transfer correlation among instances in the embedded space between for knowledge distillation.

\section{Correlation Congruence Knowledge Distillation}

In this section, we describe the details of proposed method based on correlation congruence for knowledge distillation.

\begin{figure*}[h]
    \caption{The overall framework of correlation congruence for knowledge distillation ($T$: teacher; $S$: teacher; $f_i^T$: teacher's output of $i_{th}$ sample; $f_i^S$: student's output of $i_{th}$ sample; $C_i$: correlation between $i_{th}$ and $j_{th}$ sample). Original KD focus on only instance congruence between teacher and student network. While CCKD aims to not only instance congruence but also correlation congruence between multiple instances. } \label{fig:overall_pipeline}
    \begin{center}
    \includegraphics[width=0.98\linewidth]{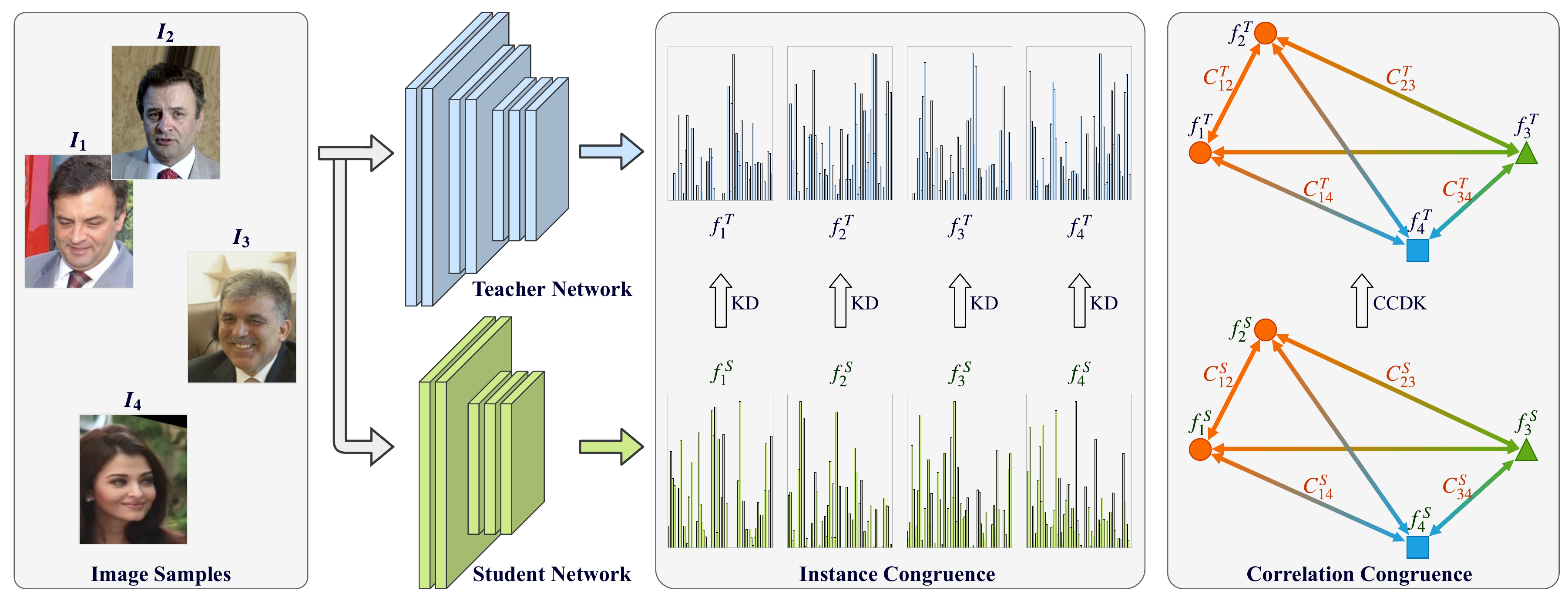}
    \end{center}
    \end{figure*}

\subsection{Background and Notations}
We refer a well-performed teacher network with parameters $\bm{W}_t$ as T and a new student network with parameters $\bm{W}_s$ as S like in \cite{hinton2015distilling,Zagoruyko2016Paying,yim2017gift,anil2018large,romero2014fitnets}. The input dataset of the network is noted as $\chi=\{\bm{x}_1,\bm{x}_2,...,\bm{x}_n\}$, and the corresponding ground truth is noted as $\mathcal{Y}=\{ \bm{y}_1,\bm{y}_2,...,\bm{y}_n \}$, $n$ represents the number of samples in dataset. Since deep network can be viewed as a mapping function stacked by multiple non-linear layers, we note $\phi_t(\bm{x};\bm{W}_t)$ and $\phi_s(\bm{x};\bm{W}_s)$ as the mapping functions of teacher and student, $\bm{x}$ represents the input data. $\bm{f}_s$ and $\bm{f}_t$ represent the feature representations of teacher and student. The logits of teacher and student note as $\bm{z}_t=\phi(\bm{x};\bm{W}_s)$ and $\bm{z}_s=\phi(\bm{x};\bm{W}_t)$. $\bm{p}_t=softmax(\bm{z}_t)$ and $\bm{p}_s=softmax(\bm{z}_s)$ represent the final prediction probilities of teacher and student. 

\subsection{Knowledge Distillation}
Overparameterized networks have shown powerful optimization properties to learn the desired mapping function from data \cite{du2018gradient}, of which the output reflects fine-grained structure one-hot labels might ignore. Based on this insight, knowledge distillation was first proposed in \cite{Bucilua2006} for model compression, then Hinton \etal \cite{hinton2015distilling} popularized it. The idea of knowledge distillation is to let the student mimic the teacher's behavior by adding a strong congruent constraint on predictions \cite{Bucilua2006,hinton2015distilling,romero2014fitnets} using KL divergence
\begin{equation}
    L_{\text{KD}} = \frac{1}{n} \sum_{i=1}^n \tau^2 KL(\bm{p}_s^{\tau},\bm{p}_t^{\tau}), \label{kd_loss}
\end{equation}
\noindent where $\tau$ is a relaxation hyperparameter (referred as temperature in \cite{hinton2015distilling}) to soften the output of teacher network, $\bm{p}^{\tau}=softmax(\frac{\bm{z}}{\tau})$. In several works \cite{Gregor2016Do,li2017mimicking} the KL divergence is replaced by euclidean distance,
\begin{equation}
    L_{\text{mimic}} = \frac{1}{n} \sum_{i=1}^n  \left \|\bm{f}_s - \bm{f}_t\right\|_2^2. \label{mimic_loss}
\end{equation}

Regardless of congruent constraint on final predictions \cite{hinton2015distilling}, feature representations \cite{Gregor2016Do} or activations of hidden layer \cite{romero2014fitnets}, these methods only focus on instance congruence while ignore the correlation between instances. Due to the gap (in capacity) between teacher and student, it's hard for student to learn a identical mapping function from teacher by instance congruence. We argue that the correlation between instances is also vital for classification since it directly reflect how the teacher model the structure of different instances in embedded feature space.

\subsection{Correlation Congruence}
In this section, we describe correlation congruence knowledge distillation (CCKD) in detail. Different from previous methods, CCKD considers not only the instance level congruence but also correlation congruence between instances. Figure \ref{fig:overall_pipeline} shows the overview of CCKD. CCKD consists of two part: instance congruence (KL divergence on predictions of teacher and student) and correlation congruence (euclidean distance on correlation of teacher and student).

Let $\bm{F}_t $ and $\bm{F}_s $ represent the set of feature representations of teacher and student respectively,
\begin{small}
\begin{equation} \label{feature_output}
    \begin{split}
        \bm{F}_t &= matrix \big{(} \bm{f}_1^t, \bm{f}_2^t,..., \bm{f}_n^t \big{)},    \\
        \bm{F}_s &= matrix \big{(} \bm{f}_1^s, \bm{f}_2^s,..., \bm{f}_n^s \big{)} .
    \end{split}
  \end{equation}
\end{small}

\noindent The feature $\bm{f}$ can be seen as a point in the embedded feature space. Without loss of generality, a mapping function is introduced as follow: 
\begin{equation} \label{correlation_matrix}
    \begin{split}
        \psi:\bm{F} \rightarrow \bm{C} \in \mathbb{R}^{n \times n}.
    \end{split}
    \end{equation}

\noindent where $\bm{C}$ is a correlation matrix. Each element in $\bm{C}$ represents the correlation between $\bm{x}_i$ and $\bm{x}_j$ in embedding space, which is defined as 
\begin{equation} \label{correlation_element}
    \begin{split}
        \bm{C}_{ij} =  \varphi(\bm{f}_i, \bm{f}_j), \quad \bm{C}_{ij} \in \mathbb{R}
    \end{split}
    \end{equation}
The function $\varphi$ can be any correlation metric, and we will introduce three metric for capturing the correlation between instances in next section. Then, the correlation congruence can be formulated as follow:
\begin{equation} \label{cc_loss}
\begin{aligned}
        L_{CC} &= \frac{1}{n^2} \left \| \psi(\bm{F}_t) - \psi(\bm{F}_s) \right \|_2^2 \\
               &= \frac{1}{n^2} \sum_{i,j} ( \varphi(\bm{f}_i^s, \bm{f}_j^s) - \varphi(\bm{f}_i^t, \bm{f}_j^t))^2. 
    \end{aligned}
    \end{equation}
 
Then, the optimization goal of CCKD is to minimize the following loss function:
\begin{equation} \label{CCKD_optimization}
    \begin{split}
        L_{CCKD} = \alpha L_{\text{CE}} + (1-\alpha) L_{\text{KD}} + \beta L_{\text{CC}} ,
    \end{split}
    \end{equation}
\noindent where $L_{\text{CE}}$ is the cross-entropy loss, $\alpha$ and $\beta$ are two hyper-paramemters for balancing correlation congruence and instance correlation.

\subsection{Generalized kernel-based correlation}
Capturing the complex correlations between instances is not easy due to a very high dimension in the embedded space \cite{ver2014discovering}. In this section, we introduce kernel trick to capture the high order correlation between instances in the feature space.

Let $\bm{x},\bm{y} \in \Omega $ represent two instances in feature space, and we introduce different mapping functions $ \mathit{k}: \Omega  \times \Omega  \mapsto R$ as correlation metric, including:

\begin{enumerate}
    \item naive MMD: $\mathit{k}(\bm{x},\bm{y})= \left | \frac{1}{n} \sum_i \bm{x}_i - \frac{1}{n} \sum_i \bm{y}_i \right | $;
    \item Bilinear Pool:  $\mathit{k}(\bm{x},\bm{y})=  \bm{x}^{\top} \cdot \bm{y}$;
    \item Gaussian RBF:  $\mathit{k}(\bm{x},\bm{y})= exp( - \frac{ \left\| \bm{x} - \bm{y} \right \|_2^2 } {2 \delta^2}) $;
\end{enumerate}

MMD can reflect the distance between mean embeddings. Bilinear Pooling \cite{lin2015bilinear} can be seen as a naive $2_{th}$ order function, of which the correlation between two instances is computed by element-wise dot product. Gaussian RBF is a common kernel function whose value depends only on the euclidean distance from the origin space.  

Comparing to naive MMD and Bilinear Pool, Gaussian RBF is more flexible and powerful in capturing the complex non-linear relationship between instances. Based on Gaussian RBF, the correlation mapping function $\phi$ can be computed by a kernel function $ \mathit{K}: \mathit{F} \times \mathit{F} \in \mathbb R^{n \times n}$, where each element can be computed as  
\begin{equation}
    {[\mathit{k}(\bm{F, F})]_{ij}} \approx \sum_{p=0}^P \alpha_p (\bm{F}_{i\cdot} \cdot {\bm{F^{\top}}_{j\cdot}})^P. \label{kernel_taylor}
\end{equation}
which can be approximated by $P$-order Taylor series. Once specifying the kernel function, then the coefficient $\alpha_p$ is also confirmed. Each element $[\mathit{k}(\bm{F},\bm{F})]_{ij}$ encodes the pairwise correlations between $i_{th}$ and $j_{th}$ features in $\bm{F}$. We take Gaussian RBF kernel function as an example, then 
\begin{equation}
    \begin{aligned}
    {[\mathit{k}(\bm{F, F})]_{ij}}&=exp(-\gamma \left \| \bm{F}_i - \bm{F}_i \right \|^2 ) \\
                       &\approx \sum_{p=0}^P exp(-2\gamma) \frac{(2 \gamma )^p}{p!} (\bm{F}_{i\cdot} \cdot {\bm{F^{\top}}_{j\cdot}})^p.  \label{kernel_taylor_approximate}
    \end{aligned}
\end{equation}
\noindent where $\gamma$ is a tunable parameter.

\subsection{Strategy for Mini-batch Sampler}

Usually, stochastic gradient descent (SGD), which samples batch of training examples uniformly at random from training dataset, is adopted to train the network and then parameters are updated using the sampled batch of examples. The naive random sampler would lead to such a situation that all examples come from different classes. Although it is an unbiased estimation to true gradient of instance congruence, uniformly sampled mini-batch gradient descent will result in a high biased estimation to gradient of intra-class correlation. To transfer the true correlation information unbiasedly, a proper sampler strategy is important.

To balance the intra-class and inter-class correlation congruence, we propose two strategies for mini-batch sampler: class-uniform random sampler (CUR-sampler) and superclass-uniorm random sampler (SUR-sampler). CUR-sampler samples by class and random selects fixed $k$ number of examples foreach sampled class (eg. each batch consists of 6 class and each class contains $k=8$ examples, forminig a 48 batch size). SUR-sampler is similar to CUR-sampler, but different in that it samples examples by superclass, a more soft form of true class generated by clustering. To get the superclass of training examples, we first extract the feature using teacher model, then use the K-means to cluster. The superclass of example is defined as the cluster it belongs to. Comparing to CUR-sampler, SUR-sampler is more flexible and tolerent for imbalance label since the superclass inflects the coarse structure of instances in embedded space.

\subsection{Complexity analysis and implementation details } 
To cope with the mini-batch training, we compute the correlation in a mini-batch. Formula \ref{kernel_taylor_approximate} involves the computation of a large pairwise matrix $b \times b$ ($b$ is the batch size), and each element is approximated by $p$-order Taylor-series with $p$ times dot product computation between two $d$ dimension vectors. The total computation complexity is $O(pbd^2)$ in a mini-batch, and the extra space comsumptation is $O(b^2+d^2)$ for storing the correlation matrix. Compared to huge parameters and computation for training deep neural network, the time and computation comsumptation for correlation congruence can be ignored. Besides, since the correlation congruent constraint is added on embedding space, it only requires that the feature dimension of student network is the same with teacher. To cope with the dismatch dimension in classification tasks, a fully-connected layer with fixed-length dimension is added for both teacher and student network, which has minor influence on other methods in this paper.

\section{Experiments}

We evaluate CCKD on multiple tasks, including image classification tasks (CIFAR-100 and ImageNet-1K) and metric learning tasks (including MSMT17 dataset ReID and MegaFace for face recognition), and compare it with closely related works. Extensive experiments and analysis are conducted to delve into the correlation congruence knowledge distillation.

\subsection{Experimental Settings}

\textbf{Network Architecture and Implementation Details} Given the steady performance and efficiency computation, ResNet \cite{he2016deep} and MobileNet \cite{sandler2018mobilenetv2} network are chosen in this work. 

In the main experiments, we set the order $P=2$, and compute Equation \ref{kernel_taylor_approximate} in a mini batch. For the networks in CIFAR-100 and ImageNet-1K, we add a fully-connected layer with 128-d output to form a sharing embedding space for teacher and student. The hyper-parameter $\alpha$ is set to zero, and correlation congruence scale $beta$ is set to 0.003, $\gamma=0.4$. CUR-sampler is used for all the main experiments with $k=4$.

On CIFar-100, ImageNet-1K and MSMT17, Original Knowledge distillation (KD) \cite{hinton2015distilling} and cross-entropy (CE) are chosen as the baselines. For face recognition, ArcFace loss \cite{Deng2018ArcFace} and $L2$-mimic loss \cite{li2017mimicking,luo2016face} are adopt. We compare CCKD with several state-of-the-art distillation related methods, including attention transfer (AT) \cite{Zagoruyko2016Paying}, deel mutual learning (DML) \cite{zhang2018deep} and conditional adversarial network (Adv) \cite{xu2018training}. For attention transfer, we add it for last two blocks as suggested in \cite{Zagoruyko2016Paying}. For adversarial training, the discriminator consists of FC($128 \times 64$) + BN + ReLU + FC ($64 \times 2$) + Sigmoid activation layers, and we adopt BinaryCrossEntropy loss to train it. All the networks and training procedures are implemented in PyTorch.

\subsection{Classification Results on CIFAR-100}

CIFAR-100 \cite{krizhevsky2009learning} consists of colored natural images with 32$\times$32 size. There are 100 classes in CIFAR-100, each class contains 500 images in training set and 100 images in validation set. We use a standard data augmentation scheme (flip/padding/random crop) that is widely used for these dataset, and normalize the input images using the channel means and standard deviations. We set the weight decay of student network to $5e-4$., batch size to 64, and use stochastic gradient descent with momentum. The starting learning rate is set as 0.1, and divided by 10 at 80, 120, 160 epochs, totally 200 epochs. Top-1 and top-5 accuracy are adopted as performance metric.

\begin{table}[h]
\begin{center}
\begin{tabular}{c|cc|cc}
\hline
\multirow{2}{*}{method} & \multicolumn{2}{l}{resnet-20} & \multicolumn{2}{l}{resnet-14} \\
                  & top-1         & top-5         & top-1         & top-5       \\
\hline\hline
CE                & 68.4          & 91.3          & 66.4          & 90.3          \\
KD                & 70.8          & 92.4          & 68.3          & 90.7          \\
DML               & 71.2          & 92.5          & 69.1          & 91.2          \\
AT                & 71.0          & 92.4          & 68.6          & 91.1          \\
Adv               & 70.5          & 92.1          & 68.1          & 90.6          \\
\hline
\textbf{CCKD} & \textbf{72.4} & \textbf{92.9} & \textbf{70.2} & \textbf{92.0} \\
\hline
\end{tabular}
\end{center}
\caption{Validation accuracy results on CIFAR-100. ResNet-110 is as teacher network, ResNet-20 and ResNet-14 as student networks. We keep the same training configuration for all the methods for fair comparasion.}
\label{tab:cifar100}
\end{table}

Table \ref{tab:cifar100} summarizes the results of CIFAR-100. CCKD gets a 72.4\% and 70.2\% of top-1 accuracy for ResNet-20 and ResNet-14, and substantially surpasses the CE by 4.0\% and 3.8\%, 1.6\% and 1.9\% over KD. For the online distillation DML \cite{zhang2018deep}, we train target network (ResNet-14 and ResNet-20) collaboratively with ResNet-110, and evalute performance of target netowrk. Comparing to other SOTA methods, CCKD still significantly  All the four distillation related methods significantly surpasses the original CE over 2\%, which verifies the effectiveness of teacher-student methods. 

\begin{figure} 
    \subfigure[KD loss and top-1 acc]{ 
      \label{fig:cifar_kl_acc} 
        \centering 
        \includegraphics[width=0.49\linewidth]{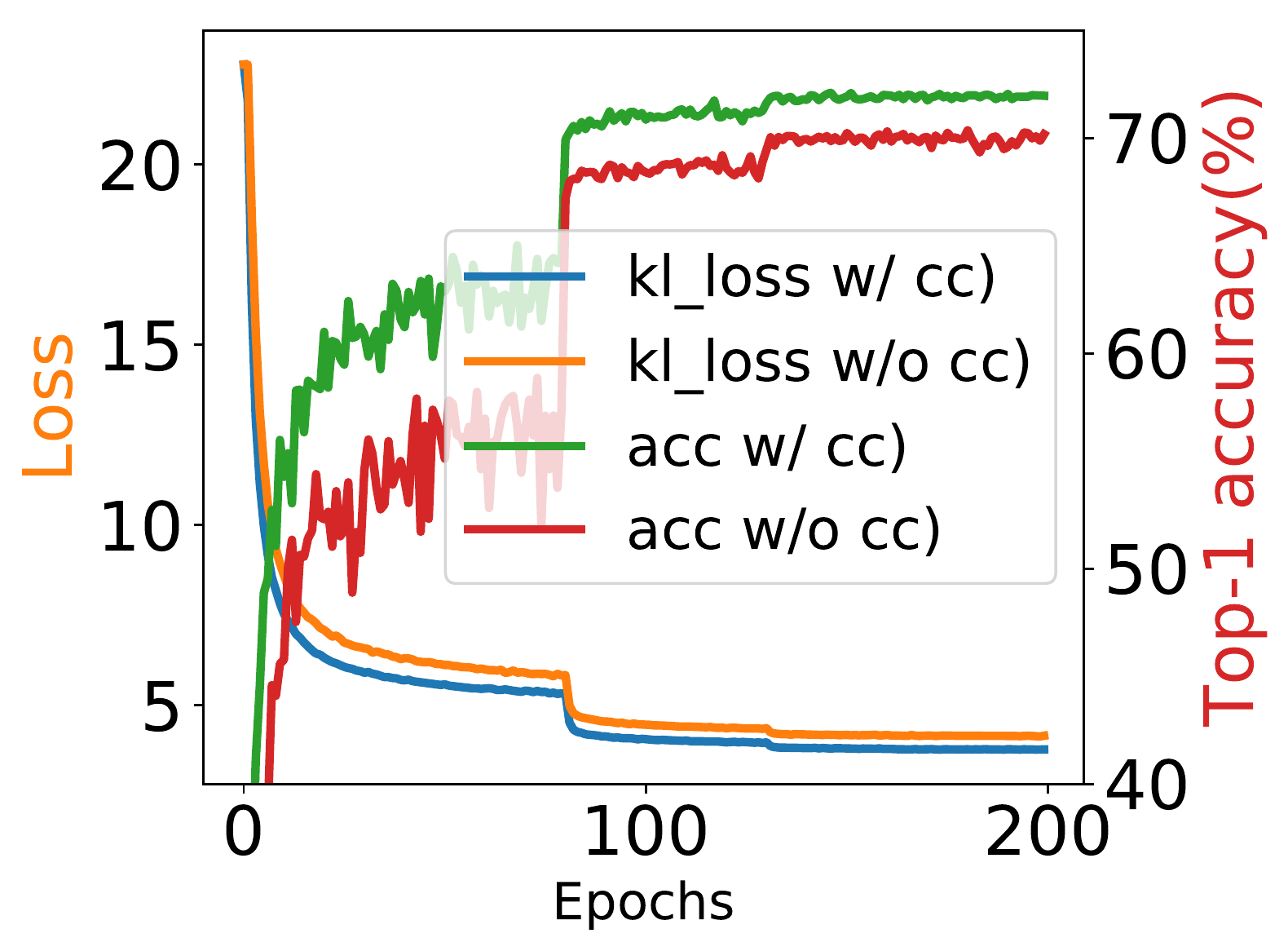}
    }
    \hspace{-2ex}
    \subfigure[CC loss]{ 
      \label{fig:cifar_cc} 
        \centering 
        \includegraphics[width=0.45\linewidth]{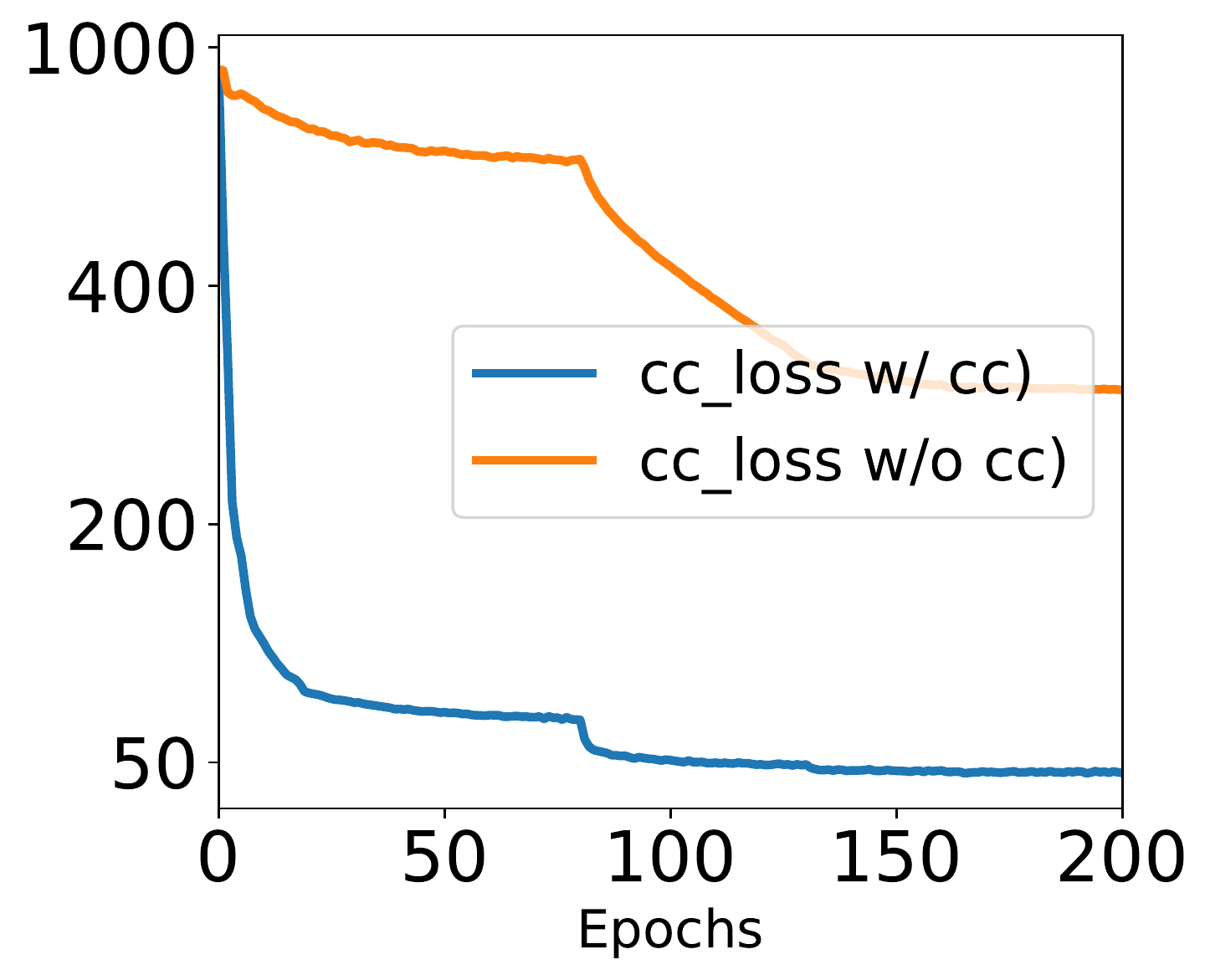}
     }
     \caption{ The curve of training loss and validation accuracy. }\label{fig:cifar_loss_acc}
   \end{figure}

Figure \ref{fig:cifar_loss_acc} shows the training loss and validation accuracy of ResNet-20. It can be observed that although KL divergence loss after convergence is almost the same, the correlation congruence loss for CCKD is much lower than original KD, consequently results in a higher performance.

\subsection{Results on ImageNet-1K}
ImageNet-1K \cite{deng2009imagenet} consists 1.28M training images and 50K testing images in total. We adopt the ResNet-50 \cite{he2016deep} as the teacher network, MobileNetV2 with 0.5 width multiplier as the student network. The data augmentation scheme for training images is the same as \cite{he2016deep}, and apply a center-crop at test time. All the images are normalized using the channel means and standard deviations. We set the weight decay of student network to $5e-4$, batch size to 1,024 (training on 16 TiTAN X, each with 64 batch size), and use stochastic gradient descent with momentum. The starting learning rate is set as 0.4, then divided  by 10 at 50, 80, 120 epochs, totally 150 epochs.

\begin{table}[h]
    \begin{center}
    \begin{tabular}{c|cc}
    \hline
    method   & top-1 accuracy & top-5 accuracy\\
    \hline\hline
    teacher  & 75.5 & 92.7 \\
    \hline
    CE       & 64.2 & 85.4 \\
    KD       & 66.7 & 87.3 \\
    DML      & 65.3 & 86.1 \\
    Adv      & 66.8 & 87.3 \\
    AT       & 65.4 & 86.1 \\
    \hline
    \textbf{CCKD}     & \textbf{67.7} & \textbf{{}87.7} \\
    \hline
    \end{tabular}
\end{center}
\caption{Validation accuracy results on ImageNet 1K. The teacher network is ResNet-50, student network is MobileNetV2 with 0.5 width multiplier. We keep the same configuration for CE and other four student networks. }
    \label{tab:imagenet_results}
\end{table}

For fair comparasion, we keep the same configuration for all the methods. Table \ref{tab:imagenet_results} summarizes the results on ImageNet 1K. CCKD gets a 67.7\% Top-1 accuracy, which surpasses the cross-entropy by promoting 3.3. Compare with original KD\cite{hinton2015distilling}, CCKD surpassses by 1.0 in top-1 accuracy. AT and DML perform worse than original KD. To our best knowledge, we have not found any works that successfully verify the effectiveness of KD on ImageNet-1K dataset. It has been reported in work \cite{Zagoruyko2016Paying} that KD struggles to work when the architecture and depth of student network are different from the teacher. But we found that by removing the dropout layer and using a proper temparature (T in [4,8]), the KD can surpass the student over 2.0\%.

\subsection{Person Re-Identification on MSMT17}
Comparing to closed set classification , open set classification is more dependent on a good metric learning and more realistic scenario. We apply the proposed method to two open-set classification: person re-identifictaion (ReID) and face recognition. 

For ReID, we evaluate proposed method on MSMT17 \cite{wei2018person}. It contains 180 hours of videos captured by 12 outdoor cameras, 3 indoor cameras under different seasons and time. There are 126,441 bounding boxes of 4,101 identities are annotated. All the bounding boxes are split to training set (32621 bounding boxes, 1041 identities), query set (11659 bounding boxes, 3060 identities) and gallery set (82161 bounding boxes). There is no intersection of identities between training set and query \& gallery set. We train the networks on training set, and perform identification on query and gallery set. Rank-1\&5 and mean accuracy precision (mAP) are adopt as performance metric.

ResNet-50 is used as the teacher network and ResNet-18 as student network. The dimension of the feature representation is set to 256. We set the weight decay to $5e-4$, batch size to 40, and use stochastic gradient descent with momentum. The learning rate is set as 0.0003, then divided  by 10 at 45, 60 epochs, totally 90 epochs.

\begin{table}[h]
\begin{tabular}{c|ccccc}
\hline
method     & pretrained?      & rank-1        & rank-5        & mAP           \\
\hline\hline
teacher     & yes     & 66.4          & 79            & 34.3          \\
\hline
CE          & no      & 32.4          & 49.0          & 14.2          \\
DML-1       & no      & 34.5          & 51.5          & 16.5          \\
DML-2       & yes     & 50.2          & 66.4          & 25.3         \\
KD          & no      & 56.8          & 72.3          & 28.3          \\
AT          & no      & 57.6          & 72.5          & 28.7          \\
Adv         & no      & 56.0          & 71.6          & 27.8          \\
\hline
\textbf{CCKD} & no   & \textbf{59.7} & \textbf{74.1} & \textbf{30.7} \\
\hline
\end{tabular}
\caption{ Validation accuracy results on MSMT17. The teacher network is ResNet-50, student network is Resnet-18. }
    \label{tab:reid_results}
\end{table}

Table \ref{tab:reid_results} summarizes the results of MSMT17 with CCKD, as well as the comparison against other SOTA methods. For fair comparasion, all the distillation based methods (except DML) are trained without ImageNet-1K pretraining. For DML, both the reuslts with/without ImageNet-1K pretraining are represented. It can be seen that the performance of the CCKD significantly surpasses KD and other SOTA KD-based methods, and promote the original KD by 3.1\% for rank-1 accuracy and 2.4\% for mAP. Without the guidance of teacher, the student trained by cross-entropy only achieves 14.2\% mAP, which is much lower than 28.3\% of KD.

\subsection{Face recognition results on Megaface}
Similar to ReID, face recognition is a classical metric learning problem. Learning a discriminative embedded space is the key to get a powerful recognition model. Usually, thousands of identities (class) are required for training a well-performed recognition model. Empirical evidence shows that mimicking the feature layer with hint-based L2 Loss can bring great improvement for small network \cite{li2017mimicking,luo2016face}. In this experiment, instead of using KD loss, we adopt the $L2$-mimic loss. MS-Celeb-1M \cite{Guo2016MS} and IMDB-Face \cite{wang2018devil} are used as training datasets. 

We choose MegaFace \cite{Kemelmachershlizerman2016The}, a very popular benchmark, as testing set to evaluate the proposed method. MegaFace aims at the evaluation of face recognition algorithms at million-scale of distractors (people who are not in the testing set). We adopt 1:N identification protocol in Megaface to evaluate the different methods. Rank-1 identification rate at different number of distractors is used as metric for evaluation. We set weight decay to 5e-4, batch size to 1024, and use stochastic gradient descent with momentum. The learning rate is set as 0.1, and divided by 10 at 50, 80, 100 epochs, 120 epochs in total. ResNet-50 is used as teacher network, and MobileNetV2 with 0.5 width multipler as student network.

\begin{table}[h]
    \begin{center}
    \scalebox{0.75}{
    \begin{tabular}{c|cccccc}
        \hline
        \multirow{2}{*}{method} & \multicolumn{6}{l}{Rank-1 Identification rate at different distractors}  \\
                   &ds=$10^1$&ds=$10^2$ &ds=$10^3$&ds=$10^4$ &ds=$10^5$ &ds=$10^6$  \\
        \hline\hline
        teacher    & 99.76 & 99.66 & 99.58 & 99.49 & 99.23 & 98.15 \\
        student    & 99.20 & 96.37 & 91.49 & 84.45 & 75.60 & 65.91 \\
        mimic      & 99.63 & 98.73 & 97.25 & 94.39 & 89.60 & 83.01 \\
        mimic+Adv  & 99.64 & 98.80 & 97.43 & 94.81 & 90.52 & 84.13 \\
        \hline
        \textbf{CCKD} & \textbf{99.66} & \textbf{99.07} & \textbf{97.93} & \textbf{95.76} & \textbf{91.99} & \textbf{86.29} \\
        \hline
        \end{tabular}
    }
    \end{center}
    \caption{ Results on Megaface. The teacher network is ResNet-50 trained on MsCeleb-1M \cite{Guo2016MS} and IMDb-face \cite{wang2018devil} using ArcFace \cite{Deng2018ArcFace}. The student network is MobileNetV2 with a width multiplier=0.5. We keep the same training configuration for mimic, mimic with Adv and CCKD. }\label{tab:megaface_results}
    \end{table}

Table \ref{tab:megaface_results} shows the results on Megaface. It can be observed that ArcFace loss, which is trained by only using pure one-hot labels, achieves 65.91\% Rank-1 identification rate with 1M distractors. When guided by the teacher using $L2$-mimic loss, the student network can achieves 83.01\%, promoting by 18.1\%. This result shows that even a much small network can get a substantial improvement of performance when designing proper target and optimization goal. By adding the constraints on correlations among instance, CCKD achieves 86.29\% Rank-1 identification rate with 1M distractors, which surpasses the mimicking by 3.28\% and 2.16\% promotion over Adv \cite{xu2018training}.

\subsection{Ablation Studies}
\textbf{Correlation Metrics.} To explore the impact of different correlation metrics on CCKD, we evaluate three popular metrics, namely max mean discrepancy (MMD), Bilinear Pool and Gaussian RBF. We approximate the Gaussian RBF by using $2$-order Taylor series.
MMD reflects the difference between two isntances in mean embeddings. Bilinear Pool evaluate the similarity of instances pair, and we adopt identity matrix as the linear matrix. When the features are normalized to unit length, it is equal to the cosine similarity. Gaussian RBF is a common kernel function whose value depends only on the euclidean distance from the origin space.

\begin{table}[h]
    \begin{center}
    \begin{tabular}{c|ccc}
        \hline
    correlation metric   & rank-1 & rank-5 & mAP   \\
    \hline\hline
    MMD      & 58.9   & 73.6   & 29.4  \\
    Bilinear & 59.2   & 73.8   & 30.2  \\
    Gaussian RBF      & 59.6   & 74.0   & 30.4  \\
    \hline
    \end{tabular}
\end{center}
\caption{ Results on MSMT17 with different correlation methods, including MMD, Bilinear Pool and Gaussian RBF. The Gaussian RBF achieves the best result.}\label{tab:correlation_metrics}
    \end{table}

Table \ref{tab:correlation_metrics} shows the results of MSMT17 with different correlation metrics. Gaussian RBF achieves the better performance comparing to MMD and Bilinear Pool, while MMD performs worst. So in the main experiments, we use the Gaussian RBF approximated by $2$-order Taylor series. All the three correlation matrics greatly surpass the original KD, which proves the effectiveness of correlation in knowledge distillation.

\textbf{Order of Taylor series.}
To exploit the high order of correlations between instances, we expand the Gaussian RBF by Tarloy series to $1$, $2$, $3$ -order respectively.

    \begin{table}[h]
        \begin{center}
        \begin{tabular}{c|ccc}
        \hline
        Expand order  & rank-1 & rank-5 & mAP  \\
        \hline\hline
        $p$=1     & 59.2   & 73.7   & 30.1 \\
        $p$=2     & 59.6   & 74     & 30.4 \\
        $p$=3     & 60.5   & 74.5   & 30.7 \\
        \hline
        \end{tabular}
    \end{center}
    \caption{ Results on MSMT17 with different order ($p=1,2,3$) Taylor series.}\label{tab:order_results}
        \end{table}

Table \ref{tab:order_results} summarizes the results on MSMT17 with approximated Gaussian RBF at different orders. It can be observed that $3$-order is better than $1,2$-order, and $1$-order performs worst. Generally speaking, expanding Gaussian RBF to high order can capture more complex correlations, and consequently achieves higher performance in knowledge distillation. 

\textbf{Impact of Different Sampler Strategies.} 
To explore a proper sampler strategy, we evaluate the impacts of different sampler strategies including uniform random sampler (UR-sampler), class-uniform random sampler (CUR-sampler) and superclass-uniform random sampler (SUR-sampler) on MSMT17 dataset. For SUR-sampler, the k-means is adopted and the number of clusters is set to 1000 to generate superclass. For fair comparasion the batchsize is set to 40 for all three strategies, and we set different $k=1,2,4,8,20$ both for CUR-sampler and SUR-sampler. 
\begin{table}[h]
    \begin{center}
    \begin{tabular}{c|cccccc}
        \hline
    sampler  & rank-1        & rank-5        & mAP           \\
    \hline\hline
    UR-sampler   & 57.2          & 72.3          & 28.6          \\
    \hline
    CUR-sampler($k$=1)     & 57.4          & 72.4          & 28.8          \\
    CUR-sampler($k$=2)     & 58.9          & 73.6          & 29.4          \\
    CUR-sampler($k$=4) & \textbf{59.7} & 74.1 & 30.2 \\
    CUR-sampler($k$=8)     & 55.7          & 71.8          & 29.1    \\
    CUR-sampler($k$=20)     & 24.7          & 40.9          & 10.7    \\
    \hline
    SUR-sampler($k$=1)     & 56.2          & 72.2          & 29.4          \\
    SUR-sampler($k$=2)     & 58.3          & 73.9          & 29.9          \\
    \textbf{SUR-sampler($k$=4)}    & 59.6          & \textbf{75.0}          & \textbf{31.1} \\
    SUR-sampler($k$=8)    & 56.2          & 72.2          & 29.4    \\
    SUR-sampler($k$=20)    & 30.1          & 47.7          & 13.7     \\
    \hline     
    \end{tabular}
\end{center}
\caption{ Results on MSMT17 with different batch sampler strategies. The teacher network is ResNet-50 and the student network is ResNet-18.}\label{tab:sampler_results}
    \end{table}

Table \ref{tab:sampler_results} summarizes the results. It can be observed that the sampler strategy have a great impact on performance. Both SUR-sampler and CUR-sampler are sensitive to the value of $k$, which plays a role of balancing the intra-class and inter-class correlation congruence. When given fixed batch size, a larger $k$ means a smaller number of classes in a mini-batch. Both CUR-sampler and SUR-sampler become worse when $k=8$ or above. A possible explanation is that small number of classes in a mini-batch results a high bias estimation for true gradient. While the SUR-sampler performs better than CUR-sampler in such bad cases. By selecting proper $k$ (eg. 2 or 4 in our experiments), Both CUR-sampler and SUR-sampler performs better than UR-sampler. 

\subsection{Analyze} 
\begin{figure}[htbp]
    \caption{The heatmaps of cosine similarities between instances pairs. The top row shows intra-class similarities and the middle row shows inter-class similarities between two identities. More intra-class heatmap are showed in bottom two rows. (best viewed in color) \label{fig:visualize_heatmap}}
    \begin{center}
    \includegraphics[width=0.95\linewidth]{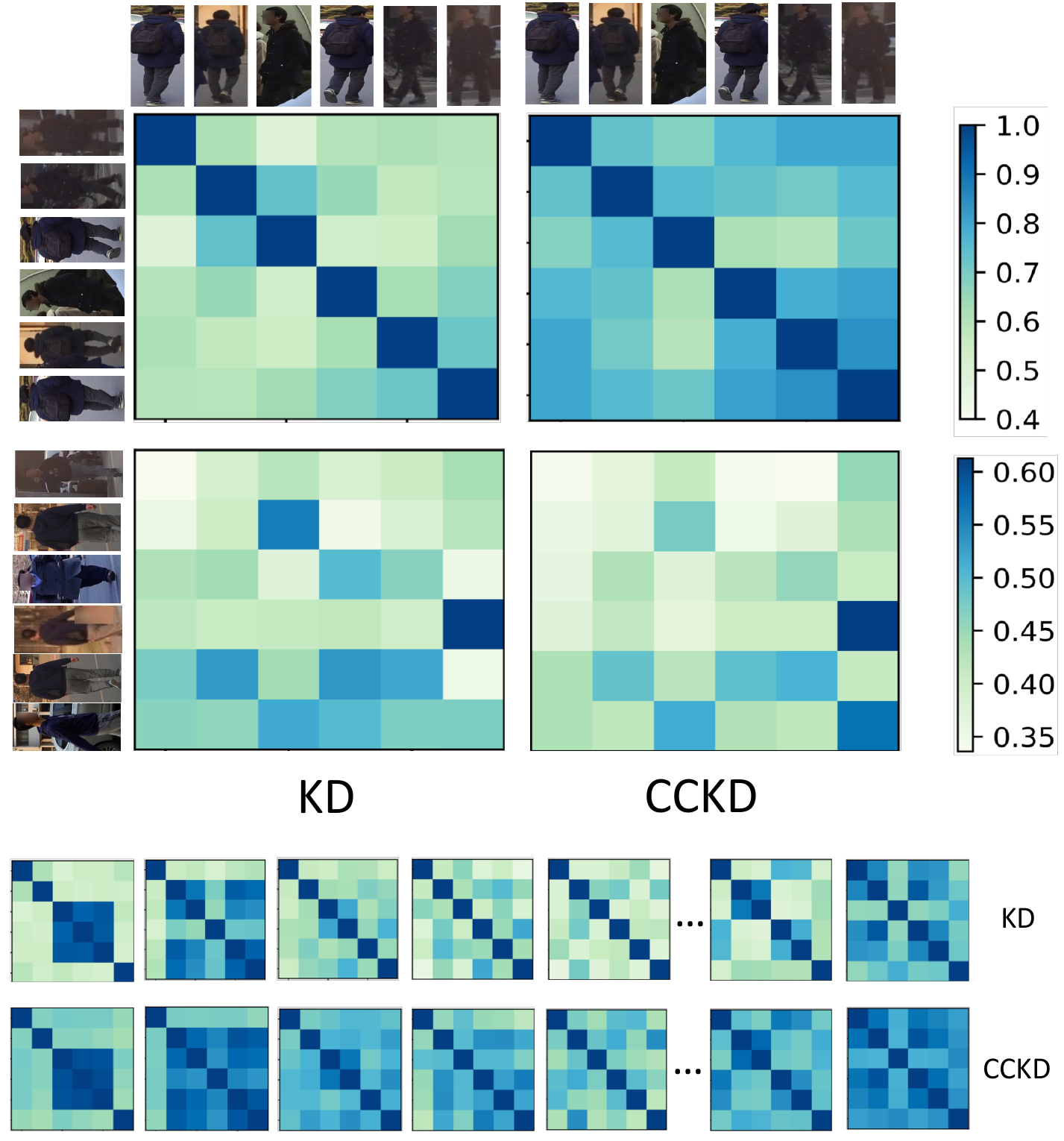}
    \end{center}
   \end{figure}

To delving into essence beyond results, we perform analysis based on visualization. We count the cosine similarities of intra-class instances and inter-class instances on MSMT17 since it is a common metric for openset recognition. Figure \ref{fig:visualize_heatmap} shows the heatmaps of cosine similarities. The top row shows intra-class instances and the bottom row shows inter-class instances from two different identities. Each cell relates to cosine similarity between correpsonding instance pair. 

It can be observed that, cosine similarity between intra-class instances of CCKD is more larger than KD overall, whichs means a more cohesion of intra-class instances in embedding space. Although there is not much difference between CCKD and KD in inter-class cosine similarity. It seems that CCKD can help the student to learn a more discriminative embedding space. While CCKD by considering the correlation congruence between instances, consequently getting a better performance.

\section{Conclusions}
In this paper, we propose a new distillation framework named correlation congruence knowledge distillation (CCKD), which considers not only instance information but also correlation information between instances when transferring knowledge. To better capture correlation, a generalized method based on Taylor series expansion of kernel function is proposed. To further improve the CCKD, two new mini-batch sampler strategies are proposed. Extensive experiments on four representative tasks show that the proposed approach can significantly promote the performance of student network.

{\small
\bibliographystyle{ieee}
\bibliography{egbib}

\begin{thebibliography}{10}\itemsep=-1pt

\bibitem{anil2018large}
R.~Anil, G.~Pereyra, A.~Passos, R.~Ormandi, G.~E. Dahl, and G.~E. Hinton.
\newblock Large scale distributed neural network training through online
  distillation.
\newblock {\em arXiv preprint arXiv:1804.03235}, 2018.

\bibitem{ba2014deep}
J.~Ba and R.~Caruana.
\newblock Do deep nets really need to be deep?
\newblock In {\em Advances in neural information processing systems}, pages
  2654--2662, 2014.

\bibitem{Bucilua2006}
C.~Bucilu\v{a}, R.~Caruana, and A.~Niculescu-Mizil.
\newblock Model compression.
\newblock In {\em Proceedings of the 12th ACM SIGKDD International Conference
  on Knowledge Discovery and Data Mining}, KDD '06, pages 535--541, New York,
  NY, USA, 2006. ACM.

\bibitem{deng2009imagenet}
J.~Deng, W.~Dong, R.~Socher, L.-J. Li, K.~Li, and L.~Fei-Fei.
\newblock Imagenet: A large-scale hierarchical image database.
\newblock In {\em Computer Vision and Pattern Recognition, 2009. CVPR 2009.
  IEEE Conference on}, pages 248--255. Ieee, 2009.

\bibitem{Deng2018ArcFace}
J.~Deng, J.~Guo, and S.~Zafeiriou.
\newblock Arcface: Additive angular margin loss for deep face recognition.
\newblock 2018.

\bibitem{denton2014exploiting}
E.~L. Denton, W.~Zaremba, J.~Bruna, Y.~LeCun, and R.~Fergus.
\newblock Exploiting linear structure within convolutional networks for
  efficient evaluation.
\newblock In {\em Advances in neural information processing systems}, pages
  1269--1277, 2014.

\bibitem{du2018gradient}
S.~S. Du, X.~Zhai, B.~Poczos, and A.~Singh.
\newblock Gradient descent provably optimizes over-parameterized neural
  networks.
\newblock {\em arXiv preprint arXiv:1810.02054}, 2018.

\bibitem{girshick2014rich}
R.~Girshick, J.~Donahue, T.~Darrell, and J.~Malik.
\newblock Rich feature hierarchies for accurate object detection and semantic
  segmentation.
\newblock In {\em Proceedings of the IEEE conference on computer vision and
  pattern recognition}, pages 580--587, 2014.

\bibitem{Guo2016MS}
Y.~Guo, L.~Zhang, Y.~Hu, X.~He, and J.~Gao.
\newblock Ms-celeb-1m: A dataset and benchmark for large-scale face
  recognition.
\newblock pages 87--102, 2016.

\bibitem{han2015deep}
S.~Han, H.~Mao, and W.~J. Dally.
\newblock Deep compression: Compressing deep neural networks with pruning,
  trained quantization and huffman coding.
\newblock {\em arXiv preprint arXiv:1510.00149}, 2015.

\bibitem{han2015learning}
S.~Han, J.~Pool, J.~Tran, and W.~Dally.
\newblock Learning both weights and connections for efficient neural network.
\newblock In {\em Advances in neural information processing systems}, pages
  1135--1143, 2015.

\bibitem{he2016deep}
K.~He, X.~Zhang, S.~Ren, and J.~Sun.
\newblock Deep residual learning for image recognition.
\newblock In {\em Proceedings of the IEEE conference on computer vision and
  pattern recognition}, pages 770--778, 2016.

\bibitem{heo2018improving}
B.~Heo, M.~Lee, S.~Yun, and J.~Y. Choi.
\newblock Improving knowledge distillation with supporting adversarial samples.
\newblock {\em arXiv preprint arXiv:1805.05532}, 2018.

\bibitem{Byeongho2018Knowledge}
B.~Heo, M.~Lee, S.~Yun, and J.~Y. Choi.
\newblock Knowledge distillation with adversarial samples supporting
  decisionboundary.
\newblock {\em arXiv preprint arXiv:1805.05532}, 2018.

\bibitem{hinton2015distilling}
G.~Hinton, O.~Vinyals, and J.~Dean.
\newblock Distilling the knowledge in a neural network.
\newblock {\em arXiv preprint arXiv:1503.02531}, 2015.

\bibitem{howard2017mobilenets}
A.~G. Howard, M.~Zhu, B.~Chen, D.~Kalenichenko, W.~Wang, T.~Weyand,
  M.~Andreetto, and H.~Adam.
\newblock Mobilenets: Efficient convolutional neural networks for mobile vision
  applications.
\newblock {\em arXiv preprint arXiv:1704.04861}, 2017.

\bibitem{Hubara2016Quantized}
I.~Hubara, M.~Courbariaux, D.~Soudry, E.~Y. Ran, and Y.~Bengio.
\newblock Quantized neural networks: Training neural networks with low
  precision weights and activations.
\newblock {\em Journal of Machine Learning Research}, 18, 2016.

\bibitem{jaderberg2014speeding}
M.~Jaderberg, A.~Vedaldi, and A.~Zisserman.
\newblock Speeding up convolutional neural networks with low rank expansions.
\newblock {\em arXiv preprint arXiv:1405.3866}, 2014.

\bibitem{Kemelmachershlizerman2016The}
I.~Kemelmachershlizerman, S.~M. Seitz, D.~Miller, and E.~Brossard.
\newblock The megaface benchmark: 1 million faces for recognition at scale.
\newblock In {\em Computer Vision and Pattern Recognition}, pages 4873--4882,
  2016.

\bibitem{krizhevsky2009learning}
A.~Krizhevsky and G.~Hinton.
\newblock Learning multiple layers of features from tiny images.
\newblock Technical report, Citeseer, 2009.

\bibitem{li2017mimicking}
Q.~Li, S.~Jin, and J.~Yan.
\newblock Mimicking very efficient network for object detection.
\newblock In {\em 2017 IEEE Conference on Computer Vision and Pattern
  Recognition (CVPR)}, pages 7341--7349. IEEE, 2017.

\bibitem{lin2015bilinear}
T.-Y. Lin, A.~RoyChowdhury, and S.~Maji.
\newblock Bilinear cnn models for fine-grained visual recognition.
\newblock In {\em Proceedings of the IEEE International Conference on Computer
  Vision}, pages 1449--1457, 2015.

\bibitem{luo2016face}
P.~Luo, Z.~Zhu, Z.~Liu, X.~Wang, X.~Tang, et~al.
\newblock Face model compression by distilling knowledge from neurons.
\newblock In {\em AAAI}, pages 3560--3566, 2016.

\bibitem{Molchanov2016Pruning}
P.~Molchanov, S.~Tyree, T.~Karras, T.~Aila, and J.~Kautz.
\newblock Pruning convolutional neural networks for resource efficient
  inference.
\newblock 2016.

\bibitem{romero2014fitnets}
A.~Romero, N.~Ballas, S.~E. Kahou, A.~Chassang, C.~Gatta, and Y.~Bengio.
\newblock Fitnets: Hints for thin deep nets.
\newblock {\em arXiv preprint arXiv:1412.6550}, 2014.

\bibitem{sandler2018mobilenetv2}
M.~Sandler, A.~Howard, M.~Zhu, A.~Zhmoginov, and L.-C. Chen.
\newblock Mobilenetv2: Inverted residuals and linear bottlenecks.
\newblock In {\em Proceedings of the IEEE Conference on Computer Vision and
  Pattern Recognition}, pages 4510--4520, 2018.

\bibitem{sau2016deep}
B.~B. Sau and V.~N. Balasubramanian.
\newblock Deep model compression: Distilling knowledge from noisy teachers.
\newblock {\em arXiv preprint arXiv:1610.09650}, 2016.

\bibitem{simonyan2014very}
K.~Simonyan and A.~Zisserman.
\newblock Very deep convolutional networks for large-scale image recognition.
\newblock {\em arXiv preprint arXiv:1409.1556}, 2014.

\bibitem{szegedy2015going}
C.~Szegedy, W.~Liu, Y.~Jia, P.~Sermanet, S.~Reed, D.~Anguelov, D.~Erhan,
  V.~Vanhoucke, and A.~Rabinovich.
\newblock Going deeper with convolutions.
\newblock In {\em Proceedings of the IEEE conference on computer vision and
  pattern recognition}, pages 1--9, 2015.

\bibitem{Gregor2016Do}
G.~Urban, K.~J. Geras, S.~E. Kahou, O.~Aslan, S.~Wang, R.~Caruana, A.~Mohamed,
  M.~Philipose, and M.~Richardson.
\newblock Do deep convolutional nets really need to be deep and convolutional?
\newblock {\em Nature}, 521, 2016.

\bibitem{ver2014discovering}
G.~Ver~Steeg and A.~Galstyan.
\newblock Discovering structure in high-dimensional data through correlation
  explanation.
\newblock In {\em Advances in Neural Information Processing Systems}, pages
  577--585, 2014.

\bibitem{wang2018devil}
F.~Wang, L.~Chen, C.~Li, S.~Huang, Y.~Chen, C.~Qian, and C.~C. Loy.
\newblock The devil of face recognition is in the noise.
\newblock {\em arXiv preprint arXiv:1807.11649}, 2018.

\bibitem{wei2018person}
L.~Wei, S.~Zhang, W.~Gao, and Q.~Tian.
\newblock Person transfer gan to bridge domain gap for person
  re-identification.
\newblock In {\em Proceedings of the IEEE Conference on Computer Vision and
  Pattern Recognition}, pages 79--88, 2018.

\bibitem{Wu2016Quantized}
J.~Wu, L.~Cong, Y.~Wang, Q.~Hu, and J.~Cheng.
\newblock Quantized convolutional neural networks for mobile devices.
\newblock In {\em Computer Vision and Pattern Recognition}, pages 4820--4828,
  2016.

\bibitem{xu2018training}
Z.~Xu, Y.-C. Hsu, and J.~Huang.
\newblock Training shallow and thin networks for acceleration via knowledge
  distillation with conditional adversarial networks.
\newblock 2018.

\bibitem{yim2017gift}
J.~Yim, D.~Joo, J.~Bae, and J.~Kim.
\newblock A gift from knowledge distillation: Fast optimization, network
  minimization and transfer learning.
\newblock In {\em The IEEE Conference on Computer Vision and Pattern
  Recognition (CVPR)}, volume~2, 2017.

\bibitem{Zagoruyko2016Paying}
S.~Zagoruyko and N.~Komodakis.
\newblock Paying more attention to attention: Improving the performance of
  convolutional neural networks via attention transfer.
\newblock 2016.

\bibitem{Zhang2017ShuffleNet}
X.~Zhang, X.~Zhou, M.~Lin, and J.~Sun.
\newblock Shufflenet: An extremely efficient convolutional neural network for
  mobile devices.
\newblock 2017.

\bibitem{zhang2018deep}
Y.~Zhang, T.~Xiang, T.~M. Hospedales, and H.~Lu.
\newblock Deep mutual learning.
\newblock In {\em Proceedings of the IEEE Conference on Computer Vision and
  Pattern Recognition}, pages 4320--4328, 2018.

\end{thebibliography}
}

\end{document}